\documentclass[10pt]{article}
\usepackage{nips2005e,times}
\usepackage{algorithmicx}
\usepackage{amssymb}
\usepackage{epstopdf}
\usepackage{graphicx}
\usepackage{graphics}

\title{On the Job Training}

\author{
Jason E. Holt \\
Computer Science Department \\
Brigham Young University\\
\texttt{isrl@lunkwill.org} \\
}

\newtheorem{theorem}{Theorem}[section]

\algblock{Setup}{End}
\algblock{Execution}{End}
\algblock{Evaluation}{End}
\algblock{Loop}{End}
\algnotext{End}

\algblock{PerfSetup}{End}
\algblock{Perf}{End}
\algblock{If}{End}
\algblock{Constants}{End}
\algnotext{End}

\algblock{ExpectedWrong}{End}

\begin{document}

\maketitle

\begin{abstract}
We propose a new framework for building and evaluating machine
learning algorithms.  We argue that many real-world problems require
an agent which must quickly learn to respond to demands, yet can
continue to perform and respond to new training throughout its useful
life.  We give a framework for how such agents can be built, describe
several metrics for evaluating them, and show that subtle changes in
system construction can significantly affect agent performance.
\end{abstract}

%
%

\section{Introduction}

This paper proposes a new way of modelling machine learning (ML),
inspired by stream-based active learning but more accurately
reflecting strategies humans employ when working with trainers and
more directly addressing the needs of users reluctant to risk effort
integrating ML algorithms.  Like active learning it aims to reduce
labeling overhead, but unlike existing models, it removes the
separation between training and testing, leading to systems which can
quickly adapt to their tasks as well as receive ongoing training
throughout their useful lives.  We call our approach ``on the job
training'' (OJT).

Traditional paradigms assume that an administrator prepares training
and test sets with distributions similar to that of the real problem,
assesses performance of the agent, then deploys a fixed instance of
the classifier once it reaches a satisfactory level of performance.
However, this train/test/use paradigm does not adequately model the
needs of systems which must begin performing immediately, or living
systems whose datasets change and for which additional classified data
becomes available over time.  The OJT framework explicitly addresses
this by formally defining a concurrent train/use process along with
performance metrics which indicate both immediate and long term
performance.  Traditional classifications distinguish between active
and passive systems, and within active learning, between pool-based
and stream-based designs.  We expand this space to include synchronous
and asynchronous relationships between agent and trainer, the presence
or absence of intervention from the trainer, and temporally evolving
distributions of interest.

This is a framework paper exploring one particular niche in a spectrum
of learning paradigms, which we claim reflects the needs of many
real-world practitioners.  Like stream-based active learning, our
model lets the agent decide how confident it feels in its
classification ability, allowing it to take over for the trainer when
easy problems arise or to ask for help when a difficult query comes
along, maximizing the value of the trainer's time.  Our performance
metrics are unique in measuring how rapidly the agent gains the
ability to relieve the trainer from tedious tasks without neglecting
long-term accuracy.  We also assume that the trainer is willing to
work interactively with the agent, giving unsolicited advice or
allowing the agent to specify what points it would most like to have
classified, as a pool-based active learner would.  These points will
often identify interesting or problematic areas for the trainer to
research.

Consider how people tend to evaluate software.  Users almost never
invest a large amount of initial effort reading a program's manual,
prefering instead to start working immediately to see if the software
does what they need.  While they have some initial patience, they
expect to see results quickly, and if the package isn't working when
their patience runs out, they'll discard it, regardless of how it
might have done in the long run.  Conversely, users also expect
software that they use over time to remember their changing
preferences, and get frustrated with software that continually
requires overrides to its initial settings.

Our metrics are designed to address these human requirements,
emphasizing the need to solve initial problems correctly with human
aid, minimizing up-front training demands, yet still requiring
satisfactory results in the long run -- in short, suitable for
on-going, real-time interaction with humans.

Consider also the problem of constructing security policies.  Low
security installations tend to use overly broad policies rather than
expend the resources required to handle the numerous exceptions and
permission grants which occur over time.  Administrators will be
unwilling to spend time with ML assistants unless they can adapt to
their task very rapidly, responding to correction, asking for help
only for boundary conditions and adapting to later policy changes.
High security installations have to balance the need for tight control
with the danger of overly complex policies which will have subtle
errors.  Their administrators are unwilling to allow an agent to make
security decisions, but can still benefit from an agent which can
point out parts of the access space which seem most unusual.  In both
cases, users want an agent which can quickly give useful results, yet
perform in the long term as conditions change.

There is little work directly related to this paper.  Active learning
traditionally focuses on different \emph{learning techniques} (such as
the relative advantages of certainty based methods \cite{lindenbaum04}
compared to committee based methods \cite{seung92}), as opposed to
exploring new \emph{performance requirements}.  Closest in spirit to
OJT are stream-based active learners \cite{saunier04}, but these have
generally been evaluated in terms of overall classifier accuracy,
rather than with respect to their ability to address the immediate
human needs we have illustrated.

In addition to defining our class of agents and corresponding
performance metrics, we will specify how to build an ideal but
impractical OJT learner, and consider an obvious approximation to that
algorithm based on KNN.  We conclude with promising performance
results and discuss future research.

%
%

\vskip -0.1in
\section{System Specification}
\vskip -0.1in

Define $S$, a sample set of zero or more unclassified points which may
be used as questions by the selective sampling algorithm (systems
which use membership queries instead of selective sampling may leave
$S$ unspecified).  $S$ is provided to the agent at the start of the
execution phase.  Define $C(q)$, a function which returns the true
classification of any point $q$. Let $G_t(q)$ denote the agent's
``guess'' at the classification of any point $q$ at a particular
``time'' $t$.  Let $V=\emptyset$, the set representing the queries
which have arrived. Let $t$ be the loop iteration counter.  Define
$Perf(t,q,Q,f,G_t(q),C(q))$, an (optional) performance metric which
keeps track of the agent's classifier throughout the execution phase.

An OJT agent executes a series of five basic steps, which may be
arranged differently depending on the system design.
The steps are 1) {\bf Query.} A query $q$ is sent to the agent. Set
$V=V \cup q$.  $q$ can be null for asynchronous agents, meaning that
no query needs to be answered at this time. 2) {\bf Question.} The
agent outputs a question $Q$, an unclassified point it wishes the
trainer to classify.  In selective sampling implementations, $Q \in S
\cup V$.  $Q=\emptyset$ implies the agent chooses not to ask a
question.  3) {\bf Fact.} The trainer sends the agent $f=\langle Q,
C(Q) \rangle$, or $f=\emptyset$ if $Q=\emptyset$. $f$ can be null for
asynchronous agents, meaning that no fact is forthcoming at this time.
Real systems might also allow the trainer to set Q, providing an
unsolicited fact.
4) {\bf Answer.} The agent outputs $G_t(q)$.  5) {\bf Assess.} Call
$Perf(t,q,Q,f,G_t(q),C(q))$ to assess the agent's performance.

Here, we make our first distinctions between agents.  A
\emph{synchronous} system requires a query to be provided at each loop
iteration, and that the trainer always answers questions asked by the
agent during the iteration in which they're asked. Note that agents
are not required to ask a question during a particular loop iteration.
Real world problems which benefit from on the job training are likely
to have an \emph{asynchronous} nature, however, in which queries may
arrive at any time, and in which trainers may not always be available
or willing to answer questions from the agent.  Furthermore, the facts
provided by the trainer are not required to correspond to the
questions asked by the agent.

The order of operations in the loop is critical, and motivates the
distinction between systems with and without \emph{intervention}.  If
the answer phase is placed after the fact phase, the agent has no
opportunity to choose its question so as to aid in answering the query
at hand, and is said to be a system without intervention.  If the
query set $V$ follows a different distribution than the sample set
$S$, then the agent can still use $V$ to specialize on part of the
problem space.  But if the distributions of $V$ and $S$ are the same,
then the agent in a system without intervention has no opportunity to
improve its accuracy for an unanswered query.

An agent with intervention therefore executes the steps in the
following order: \textbf{Query, Question, Fact, Answer, Assess}.  An
agent without intervention executes the steps in this order:
\textbf{Query, Answer, Question, Fact, Assess}.  Synchronous and
asynchronous systems are distinguished by whether or not null queries
and facts are permitted.

%
%

\vskip -0.1in
\section{Performance Metrics}
\vskip -0.1in

Since OJT aims to encourage creation of both agents which respond
quickly to new tasks and agents which are used and trained over long
periods of time, we propose two new performance metrics.  The first
fixes all but two variables, giving the algorithm a fixed budget of
questions to ask.  The second metric uses an arbitrary utility
function, allowing question and wrong answer values to be externally
provided.  A crucial point is that these metrics are made available to
the agent, which allows it to optimize with respect to the performance
measure it is being evaluated by.

Both metrics calculate agent accuracy cumulatively, {\it as the
queries are answered}.  Because of the lack of a formal testing phase,
we calculate accuracy using the answers provided at each step. The
focus on accuracy from the start requires a finite limit on the number
of rounds considered if there is also to be a finite limit to the
number of questions.  Otherwise, as we will show later in this
section, cumulative accuracy becomes indistinguishable from
traditional accuracy measurements.

For the first metric, this requires the user to decide how important
the agent's learning rate is as a function of the amount of training
he is willing to provide.  This is natural decision -- a user trying to
solve a problem in one hour wants an algorithm which will very quickly
be of some assistance, while a user preparing an agent for high volume,
long term service will be willing to offer more training to an algorithm
in exchange for high overall accuracy.

{\bf Budget Metric.\ \ } Our first metric mirrors the traditional
active learning model in which an algorithm chooses a preset number of
points for classification.  Since this metric's utility lies in the
simplicity of choosing only question and query limits, we specify it
with respect to a synchronous system (but it is readily adapted to
other variants). The metric computes the agent's cumulative error over
the course of execution.  Agents are expected to use all available
questions; no reward is given for unused questions at termination.

Define $k_q$ to be the query limit and $k_Q$ to be the question
budget.  Let $c$ be the counter for incorrectly classified queries,
and $Q_{rem}=k_Q$ be the number of questions remaining.  The budget
metric is then defined as the {\bf cumulative error} rate, or $c/k_q$
(when $t=k_q$).

This simple measure would be no different from traditional error
calculations except that $c$ is accumulated after each query is
answered, in the {\bf Assess} phase.  In section \ref{intervention} we
give two theorems which demonstrate the significance of this seemingly
small difference.  Note that, since this metric considers the
\emph{rate} at which each learner learns, our metric is quite useful
for evaluating active learners, since their strength is generally
considered to be in their fast convergence.

{\bf Utility metric.\ \ } In the utility metric, the {\bf Assess}
phase calculates $c$, the cumulative cost as $\sigma C_Q(Q_t) +
C_w(G_t(q))$, where $C_Q$ and $C_w$ denote the cost of asking a
particular question or being wrong about a particular query.  If
possible, these cost functions should be made available to the agent.
At the risk of being too general, this metric illustrates that agents
should take advantage of questions that are easy for a trainer to
answer (or times when the trainer is available to work with the agent)
and should be willing to guess when a wrong answer will not be
expensive.

%
%

\vskip -0.1in
\section{Intervention and $k_q$}\label{intervention}
\vskip -0.1in

In this section, we show how two of the seemingly minor elements of
our system and metric definitions have a dramatic impact on agent
strategy and performance.  First, we show that systems without
intervention have advantages over traditional active learners only
when the distribution of queries over time reveals information about
the task at hand.  Second, we emphasize that OJT's distinctions from
traditional active learning fade once questions are no longer being
answered, and that consequently, the choice of $k_q$ in our proposed
metrics is critical in making useful assessments of an OJT system.

\begin{theorem}
Assuming queries are chosen uniformly at random, with replacement,
from a set $T$ whose distribution is the same as a sufficiently large
unlabeled sample set $S$, an agent in a synchronous OJT system without
intervention has no statistical advantage over a traditional active
learning agent with respect to the budget metric.
\end{theorem}

{\bf Proof:} Consider the first iteration for an agent $A$. $A$ is
forced to answer a query $q$ before receiving the classification for
any point.  Once it provides its (random) guess, it is allowed to
choose any point as its question.  Since the trainer returns a fact
only after the agent has output its guess, the fact clearly cannot
influence that guess.  But since $q$ was chosen randomly from $T$, and
$T$ has the same distribution as $S$, it provides no information, in a
statistical sense, about any future query.  Consequently the agent can
discard $q$ after guessing its classification.  But this, again,
happens before the question phase, so $q$ provides no useful
information about what question to ask.  By induction, we see that the
agent has the same dilemma for all future iterations of the loop.  The
agent can therefore equivalently output its classifier $G_t$ in each
iteration of the loop before the query arrives, then ignore the query
once it does, since the query provides no utility to the agent for
current or future rounds.  Thus $A$ can equivalently operate as either
an OJT agent or a traditional active learner with respect to the
budget metric. $\Box$

In the event that $T$ and $S$ have different distributions, OJT agents
have an advantage over active learners even in systems without
intervention.  But also note that this proof doesn't apply to systems
with intervention, in which OJT can outperform baseline even for
problems which are traditionally considered unlearnable.

\vskip -0.1in
\subsection{The importance of $k_q$}
\vskip -0.1in

While OJT systems should generally produce agents which have good
overall accuracy, their emphasis is on learning the query set at hand.
While this seems natural for many applications, it is useful to note
that if all you want is a system to train once and then deploy, OJT
has little to offer over traditional active learning, as this theorem
demonstrates.

\begin{theorem}
Define an optimal agent as one whose classifier accuracy monotonically
increases as it learns queries and facts, and which achieves
cumulative error less than or equal to the cumulative error of any
other practical classifier. Then given a finite $k_Q$ and an infinite
query set $T$, the difference in cumulative error $\epsilon$ between
optimal agent A in an OJT system with and intervention and optimal
agent B in an OJT system without intervention approaches zero as $k_q$
approaches infinity.
\end{theorem}

{\bf Proof:} Assume that an optimal agent $A$ in an OJT system with
intervention can achieve a cumulative error of $e_A$ by asking no more
than $k_Q$ questions over $k_q$ queries from $T$.  An optimal agent
$B$ in a system without intervention can receive precisely the same
information available to $A$, but only at a later time, so its
cumulative error $e_B$ can never be lower than $e_A$.  Let
$\epsilon=e_B-e_A$.  Assume that $\epsilon$ is maximized, so that $A$
answers $k_Q$ queries correctly which $B$ misclassifies as a result of
not receiving facts until later in its loop.  Then for a given
$k_q>k_Q$, $\epsilon = k_Q/k_q$, which approaches 0 as $k_q$
approaches infinity. $\Box$

Despite our questionable definition of an ``optimal agent,'' it should
be clear that the advantages of intervention fade as one considers
cumulative accuracy past the point where the question budget has been
exhausted.  This should not be taken to mean that OJT is no different
from other types of ML; rather, it should emphasize that OJT's
strengths lie in adaptability to temporally evolving conditions and
ability to minimize load on trainers.

%
%

\vskip -0.1in
\section{Implementation}
\vskip -0.1in

Here we show how to create both an ideal, but impractical OJT agent as
well as a practical agent which generally outperforms its active
learning counterpart.  Both implementations assume a budget metric and
synchronous systems with intervention.

Modeling OJT agent strategies is easier if the agent can predict how
accurate it will become if given the classification of a particular
unlabeled point.  The implementations we propose all assume that the
underlying classifier has the ability to recursively predict how it
will classify future points, and how confident it will be in that
classification.  Let such an algorithm provide three functions:
$Unc(p)$ returns the probability that p will be misclassified.
$Add(p)$ assumes that $p$'s label is known when calculating
$Unc(\cdot)$ ($Add$ may be called more than once to assume multiple
points are known). $Remove(p)$ means that the algorithm should no
longer assume that $p$'s label is known.

\vskip -0.1in
\subsection{Ideal OJT}
\vskip -0.1in

Given an agent with perfect foresight and unlimited computing power,
and some simplifying assumptions about the system, it is
straightforward to construct an ideal system in the sense that it
minimizes cumulative error with respect to the budget metric.  We
begin by assuming the entire test set $T$ and remaining test set $R$
are known, and that queries are chosen from $R$ uniformly at random.
Let $q$ be the query just selected in the query phase (which is no
longer in $R$), and $Q_{rem}$ be the number of questions remaining in
the question budget.  Let $Permute(T,f)$ calculate all permutations of
the elements of $T$, calling $f$ on each permutation; let $Average(S)$
returns the average of the elements in set $S$.  Then the ideal OJT
algorithm for a selective sampling synchronous OJT system with
intervention is as follows.  Define a function $ExpectedWrong$, which
returns the number of queries it expects to misclassify by the time
$R$ is exhausted (which is minimal on average given our assumptions
and its choice of $Q$).  Assume $ExpectedWrong$ is called during the
question phase of system execution, and that the value $Q$ is the
question it selects (we omit the corresponding algorithm which
optimizes the utility metric rather than the budget metric, which is
easily derived).  Letting the function {\bf AvgPenalty}$(R,Q_{rem})$
return $Average(Permute(R, ExpectedWrong(R[0], R \setminus R[0],
Q_{rem})))$:

\vskip 0.1in
\begin{algorithmic}
\ExpectedWrong$(q,R,Q_{rem})$:
\State {\bf If} $q==R==\emptyset$: Set $Q=\emptyset$; Return $0$.

\State // Start assuming we ask no question
\State  Let $Q = QBest = \emptyset$
\State  Let $MinPenalty = Unc(q) + AvgPenalty(R,Q_{rem})$

\State // Then consider all possible questions we might ask
\State {\bf If} $Q_{rem}==0$, Return $MinPenalty$
\Loop ~for all $Q \in (S \cup T)$:

\State	Set $Penalty = Unc(q) + AvgPenalty(R,Q_{rem}-1)$

\If$(MinPenalty > Penalty)$
\State	  Set $MinPenalty = Penalty$
\State	  Set $QBest = Q$
\End

\End

\State  Set $Q=QBest$; Return $MinPenalty$

\End
\end{algorithmic}

Note that since we assume questions are chosen randomly from $R$, we
can average the penalties over all possible futures, choosing the
question in this round which leads to an optimal outcome on average,
since all sequences are equally likely.

Note that this algorithm is entirely intractable; $Permute$ calls
$ExpectedWrong$ for every possible future, and $ExpectedWrong$ is
itself a recursive function.  But it does suggest a general form which
a practical algorithm might take if it can approximate or efficiently
calculate the expected penalty across reasonable futures.

\vskip -0.1in
\subsection{A Tractable Approximation}
\vskip -0.1in

The KNN classifier described in \cite{lindenbaum04} provides an
excellent starting point for implementing OJT classifiers.  Their
classifier implements active learning using selective sampling, and
outperforms similar passive learners as well as several active
learning approaches adapted from other sources; it also makes it
simple to construct the $Add$, $Remove$ and $Unc$ functions required
by our implementation.  In fact, conditional uncertainty with
lookahead is the basis for their utility metric.

Our first approximation of the ideal algorithm effectively answers the
question, ``if I have only one question to ask, and I ask it now, what
should it be?''  It approximates this implementation of
$ExpectedWrong$ by substituting $S \cup V$ as an approximation to $R$.
Using $Unc(q|Q)$ as a shorthand form of $Add(Q); Unc(q); Remove(Q)$,
the algorithm is:

$$\min_{Q \in S \cup V} Unc(q|Q) + \frac{q_{rem}-Q_{rem}}{|R|} \sum_{p \in R} Unc(p|Q)$$

The summation of uncertainties divided by $|R|$ gives the expected
overall classifier error given the true classification of the $Q$
under consideration.  Multiplying by $q_{rem}$ would give the expected
number of misclassifications given no further questions, but first we
subtract $Q_{rem}$, since at least that many additional queries can be
classified correctly just by passing them along as questions.  We then
add the penalty of misclassifying the current query explicitly, since
a penalty will certainly be incurred if it is misclassified, as
opposed to the possible future queries whose uncertainty is amortized.
Most obviously missing from this implementation is the option of
asking no question at all in a given round, since we have not yet
found a reliable, efficient way to predict when this behavior will be
worthwhile.


\begin{figure}
\centering
\includegraphics[scale=0.55]{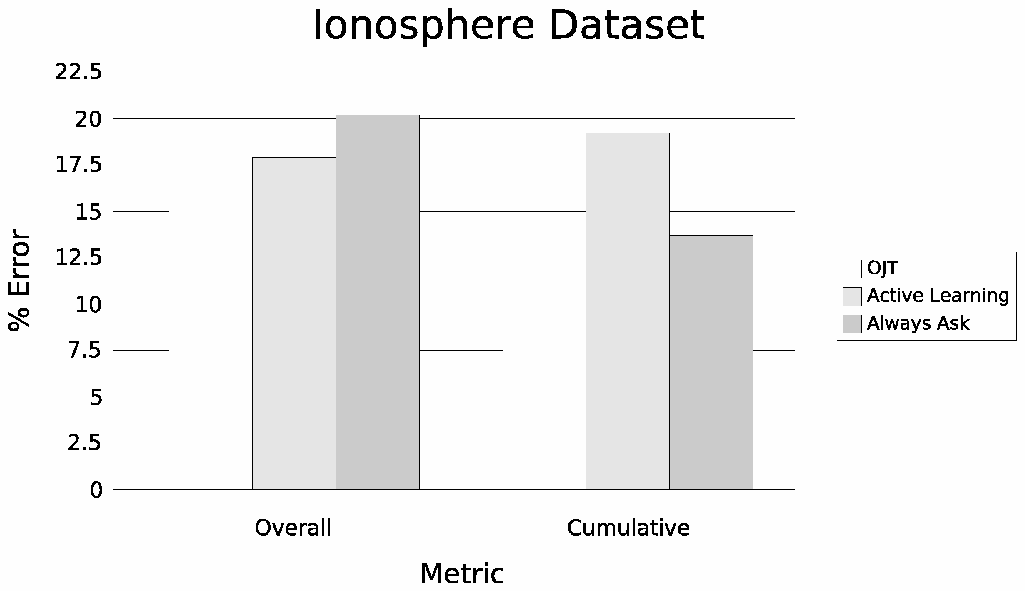}
\includegraphics[scale=0.55]{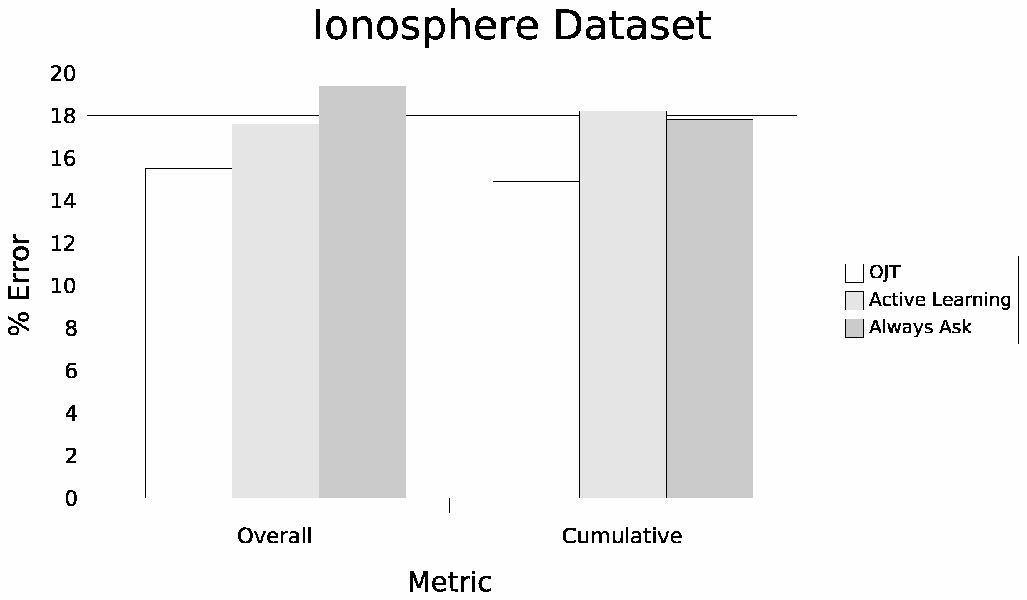}
\caption{On the left, $k_Q=20 ,k_q=50$. With 2.5 times more queries
than questions, the OJT agent's advantage in both cumulative and
overall error is sharply defined.  The ``always ask'' algorithm
suffers from its poor overall classifier, while the activer learner
cannot take advantage of the first 20 queries.  On the right, $k_Q=20
,k_q=100$.  With 5 times more queries than questions, we see
cumulative error beginning to converge with overall error, as our
second theorem predicts.  OJT still wins in the cumulative metric,
although the active learner closes the gap in the overall metric, and
in an independent overall metric would be performing just as well as
the OJT agent (see text). All results averaged over 200 runs. }
\end{figure}

\begin{figure}
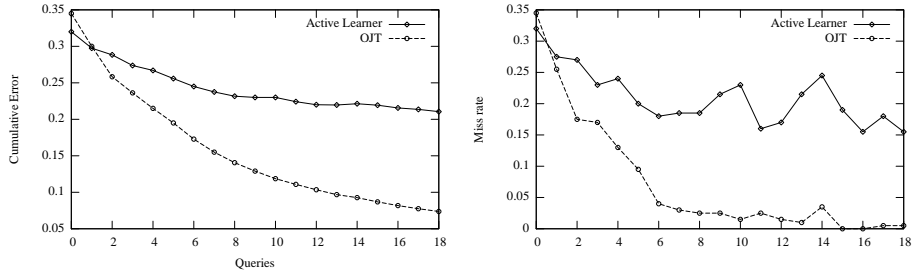

\centering
\scalebox{0.48}{\input{cumulative_error.tex}}
\scalebox{0.48}{\input{missrate.tex}}
\caption{ Performance results on the Ionosphere dataset. On the left,
 $k_Q=20 ,k_q=100$. Here we see how the active learner performs for
 the first $k_Q$ queries, gradually converging toward its overall
 accuracy.  The OJT agent, however, quickly becomes quite good at the
 query set, since it frequently chooses the queries themselves as
 questions for the trainer.  Once it runs out of questions the
 cumulative error rate will gradually {\it increase} as it approaches
 its overall accuracy threshold.  On the right, $k_Q=20
 ,k_q=100$. This graph shows the rate at which each classifier misses
 each of the first $k_Q$ queries.  Here we can see even more
 strikingly the advantage of the OJT agent in adapting to queries
 while still building a good overall classifier. }
\end{figure}

\begin{table}
\centering
\begin{tabular}{ccc|ccc}
Dataset & Metric & $k_q$ & OJT & Active & Always Ask \\
\hline
Ionosphere   & Cumulative/Overall & 50 & 11.0/15.3 & 19.2/17.9 & 13.7/20.2 \\
Ionosphere   & Cumulative/Overall & 100 & 14.9/15.5 & 18.2/17.6 & 17.8/19.4 \\

Segmentation & Cumulative/Overall & 50 & 8.7/11.8  & 13.1/11.5 & 9.4/13.6  \\
Segmentation & Cumulative/Overall & 100 & 11.7/11.4 & 12.2/11.0 & 12.6/14.4 \\

Pima Indians & Cumulative/Overall & 50 & 25.9/34.1 & 35.2/35.1 & 20.1/33.9  \\
Pima Indians & Cumulative/Overall & 100 & 34.3/35.3 & 35.0/35.0 & 28.7/34.5 \\

\end{tabular}
\caption{Error rates averaged over 100 runs}
\end{table}

\vskip -0.1in
\subsection{Performance}
\vskip -0.1in

We compared our classifier to two others.  The first uses the simple
strategy of using the first $k_Q$ queries as questions as they arrive.
This ensures 0\% cumulative error for the first $k_Q$ rounds and
performs as a passive learner would for the remaining queries.  The
second strategy is a simple adaptation of the active learning
algorithm proposed in \cite{lindenbaum04}.  It adds queries to its
unlabeled set as they come in, but otherwise behaves normally.  In all
cases, we used the budget metric and an overall accuracy metric with
$k_Q=20$ for evaluation purposes, since the results in
\cite{lindenbaum04} used approximately 20 training examples for each
classifier they tested and because the budget metric is easier to
directly compare with overall accuracy metrics.  Note that we were
unable to reproduce the exact results in \cite{lindenbaum04}.

Figure 1 shows our results on the ionosphere dataset from UCI
\cite{uci}.  The active learning algorithm's representation in the
overall metric is not entirely fair, since the overall metric uses the
same queries sent to the agents during execution.  An entirely
separate test set would give a better picture of the true generalized
accuracy of each classifier, although our metric is a good halfway
point between such a general metric and our cumulative metric.  But as
our second theorem points out, as $k_q$ increases, such advantages
disappear anyway.  This can be seen on the right graph, where the
cumulative metric more closely matches the overall metric and the gap
between the active and OJT learners decreases slightly.  We expect
that gap to disappear entirely given a large enough $k_q$.  
Correcting for this inequity analytically, we see that the results are
also consistent with what we expect.  Given the active learner's
17.6\% overall accuracy, we would expect it to be wrong on about 3.5
of the first 20 queries, while the OJT agent might get all 20 correct.
The test set used had 150 elements, so we expect the OJT learner's
results to be about 2.3\% higher than they would be on a completely
independent test set would indicate.  Since this is almost exactly the
gap between the active and OJT learners, we suspect such a test would
show that their long-term accuracies are almost identical.

Figure 2 shows how the OJT agent quickly adapts to the query set,
achieving a low cumulative error rates by the time the question budget
is exhausted.  Finally, Table 1 lists our results from two other
datasets.  The Image Segmentation dataset produced results comparable
to the Ionosphere set, although the active learner edged out the OJT
agent even with the handicap discussed earlier.  The Pima Indians
database was surprising, however, in that the ``always ask'' strategy
outperformed both other agents in both cumulative and overall
accuracy; investigating this result is left for future research.

%
%

\vskip -0.1in
\section{Conclusions}
\vskip -0.1in

This is a wide-open research area.  Our practical implementation
provides an initial attempt at maximizing the goals of OJT, and can be
vastly improved.  Asynchronous systems combined with creative cost
functions in the utility metric provide a host of possible directions
for further research, including consideration of costs which vary over
time, and burstiness of fact input when trainers are only
intermittently available in asynchronous systems.

%
%

%

%
%

{\small
\bibliographystyle{plain}
\bibliography{ojt}
}
\end{document}